\documentclass[10pt,twocolumn,letterpaper]{article}

\usepackage[pagenumbers]{cvpr} 
\usepackage{multirow}
\usepackage{amsmath}
\usepackage{algorithm}
\usepackage{algorithmic}
\usepackage{booktabs}
\usepackage{microtype}
\usepackage{bibentry}
\usepackage{booktabs}
\usepackage{amsmath} 
\usepackage{subcaption}
\usepackage{multirow}
\usepackage{makecell}
\usepackage[dvipsnames,table]{xcolor}
\usepackage{amssymb}
\usepackage{pifont}
\usepackage{graphicx}
\usepackage{subcaption}

\definecolor{blue1}{HTML}{196ab1}
\definecolor{blue2}{HTML}{4886c1}
\definecolor{blue3}{HTML}{5e9bd6}
\definecolor{blue4}{HTML}{77b1e2}
\definecolor{blue5}{HTML}{bdd930}
\definecolor{blue6}{HTML}{dfebf6}

\definecolor{red1}{HTML}{de512c}
\definecolor{red2}{HTML}{f2642d}
\definecolor{red3}{HTML}{f68f58}
\definecolor{red4}{HTML}{febf92}
\definecolor{red5}{HTML}{f8e9c8}

\definecolor{lightblue}{RGB}{225,225,225}  
\definecolor{cvprblue}{rgb}{0.21,0.49,0.74}
\usepackage[pagebackref,breaklinks,colorlinks,allcolors=cvprblue]{hyperref}

\def\paperID{2220} 
\def\confName{CVPR}
\def\confYear{2026}

\makeatletter
\def\@fnsymbol#1{\ensuremath{\ifcase#1\or \dagger\or \ddagger\or
   \mathsection\or \mathparagraph\or \|\or **\or \dagger\dagger
   \or \ddagger\ddagger \else\@ctrerr\fi}}
\makeatother

\newcommand{\homepage}{\raisebox{-1.5pt}{\includegraphics[height=1em]{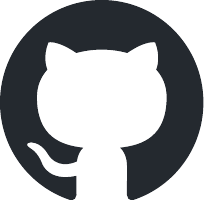}}}
\newcommand{\hfmodel}{\raisebox{-1.5pt}{\includegraphics[height=1em]{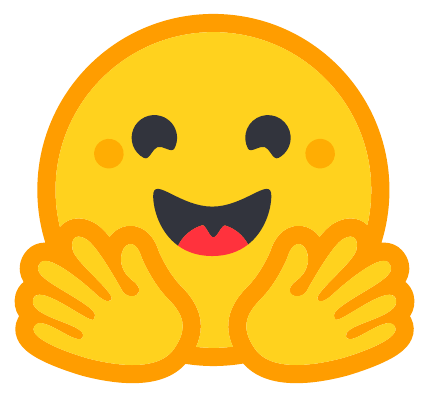}}}

\title{AdaTooler-V: Adaptive Tool-Use for Images and Videos}

\author{
\begin{tabular}{c}
\textbf{Chaoyang Wang}$^{1,6}$
\quad\!\!\!\!
\textbf{Kaituo Feng}$^{1}$\thanks{Project Leader.}
\quad\!\!\!\!
\textbf{Dongyang Chen}$^{2}$
\quad\!\!\!\!
\textbf{Zhongyu Wang}$^{3}$
\quad\!\!\!\!
\textbf{Zhixun Li}$^{4}$ \\[1ex]
\textbf{Sicheng Gao}$^{7}$
\quad\!\!\!\!
\textbf{Meng Meng}$^{6}$\thanks{Corresponding Authors.}
\quad\!\!\!\!
\textbf{Xu Zhou}$^{6}$
\quad\!\!\!\!
\textbf{Manyuan Zhang}$^{1}$
\quad\!\!\!\!
\textbf{Yuzhang Shang}$^{5}$
\quad\!\!\!\!
\textbf{Xiangyu Yue}$^{1}$\footnotemark[2] \\[1ex]
\normalfont
$^{1}$MMLab, CUHK
\quad
$^{2}$THU
\quad
$^{3}$SJTU
\quad
$^{4}$DB Group, CUHK
\quad
$^{5}$UCF
\quad
$^{6}$Sangfor
\quad
$^{7}$JMU\\[1ex]
{\homepage\ \normalfont 
\texttt{Home: \!\!\!\!\!\url{https://github.com/CYWang735/AdaTooler-V}}} \\
{\hfmodel\ \normalfont \texttt{HF: \!\!\!\url{https://huggingface.co/AdaTooler-V}}} \\
\end{tabular}
}

\begin{document}
\maketitle
\begin{abstract}
Recent advances have shown that multimodal large language models (MLLMs) benefit from  multimodal interleaved chain-of-thought (CoT) with vision tool interactions. 
However, existing open-source models often exhibit blind tool-use reasoning patterns, invoking vision tools even when they are unnecessary, which significantly increases inference overhead and degrades model performance.
To this end, we propose AdaTooler-V, an MLLM that performs adaptive tool-use by determining whether a visual problem truly requires tools. 
First, we introduce AT-GRPO, a reinforcement learning algorithm that adaptively adjusts reward scales based on the Tool Benefit Score of each sample, encouraging the model to invoke tools only when they provide genuine improvements.
Moreover, we construct two datasets to support training: AdaTooler-V-CoT-100k for SFT cold start and AdaTooler-V-300k for RL with verifiable rewards across single-image, multi-image, and video data.
Experiments across twelve benchmarks demonstrate the strong reasoning capability of AdaTooler-V, outperforming existing methods in diverse visual reasoning tasks. Notably, AdaTooler-V-7B achieves an accuracy of 89.8\% on the high-resolution benchmark V*, surpassing the
commercial proprietary model GPT-4o and Gemini 1.5 Pro.
All code, models, and data are released.

\end{abstract}

\section{Introduction}

\begin{figure}[t!]
  \centering
  \includegraphics[width=\linewidth]{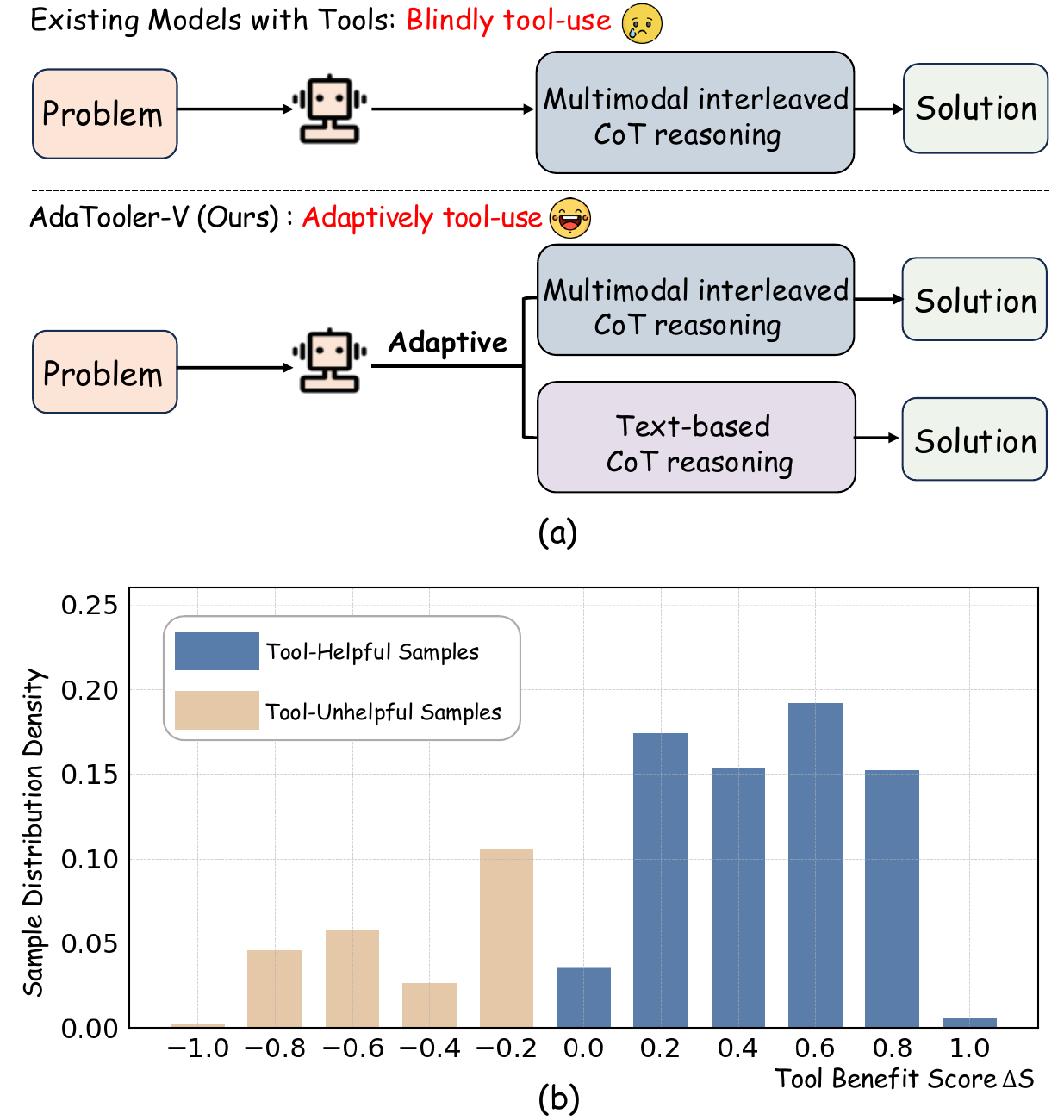}
  \caption{(a) Compared with existing models that blindly invoke vision tools, AdaTooler-V adaptively invokes tools by determining whether the problem truly requires tools. (b) Distribution of $\Delta S$ values in the AdaTooler-V-300k dataset, where positive and negative values correspond to tool-helpful and tool-unhelpful samples. Here, $\Delta S$ is computed as the difference in average accuracy when Qwen2.5-VL-72B-Instruct \cite{bai2025qwen2} solves the same sample with and without tool-use.}
  \label{fig:motivation}
\vspace{-0.1in}
\end{figure}

Recent advancements have highlighted the potential of rule-based Reinforcement Learning (RL) in enhancing the reasoning abilities of Large Language Models (LLMs) \cite{deepseekr1, yu2025dapoopensourcellmreinforcement,zhang2025critique}.
In particular, DeepSeek-R1 \cite{deepseek_r1} demonstrates the effectiveness of employing the GRPO \cite{deepseekmath} algorithm to incentivize strong reasoning  with long Chain-of-Thought (CoT) in LLMs.
Inspired by DeepSeek-R1’s success, many subsequent studies have extended this paradigm to Multimodal Large Language Models (MLLMs) \cite{huang2025visionr1incentivizingreasoningcapability, tan2025reasonrftreinforcementfinetuningvisual, wang2025vision, feng2025videor1, fan2025sophiavl}.
Notable examples include Vision-R1 \cite{huang2025visionr1incentivizingreasoningcapability}, Video-R1 \cite{feng2025videor1} and OneThinker \cite{feng2025onethinker}, which apply RL to improve visual reasoning abilities.

In the field of multimodal reasoning, a rising trend is the multimodal interleaved CoT paradigm, also known as “Thinking with Images.” In this paradigm, models dynamically interact with external vision tools (e.g., cropping, frame extraction) throughout the reasoning process \cite{zheng2025deepeyesincentivizingthinkingimages, lai2025minio3scalingreasoningpatterns, wang2025pixelreasonerincentivizingpixelspace, zhang2025thymethinkimages}. Such visual interactions enable the model to repeatedly focus on fine-grained visual details that text-only reasoning would otherwise overlook, thereby yielding substantial performance gains on challenging visual tasks.

However, existing models usually exhibit blind tool-use, invoking vision tools even when they are unnecessary. This phenomenon stems from a limitation in current approaches: models lack an explicit mechanism for determining when tools should be invoked, and reward functions may even blindly encourage tool-use.
Nevertheless, as illustrated in Fig. \ref{fig:motivation}, not all problems require tool-use.
Many visual reasoning tasks can be solved efficiently using text-based CoT, and forcing tool-use can even degrade the final prediction quality.
This is primarily because blind tool-use can induce overthinking \cite{fan2025missingpremiseexacerbatesoverthinking,chen2025think23overthinkingo1like} during reasoning, driving the model to explore unnecessary trajectories, and deviate from the optimal reasoning path \cite{su2025underthinkingoverthinkingempiricalstudy}. 
Moreover, frequent and unnecessary tool invocations gradually weaken the model’s reliance on the original visual input, making it harder for the model to focus on critical visual cues \cite{tian2025thoughtaccuracydualnature}. In addition, blind tool-use may induce a series of meaningless tool operations \cite{li2025adaptivetooluselarge}. For tasks that inherently do not require tool-use, each extra tool-use introduces unnecessary computational overhead, thereby increasing the overall inference cost.

To address these challenges, we propose AdaTooler-V, an MLLM equipped with adaptive tool-use ability.
Unlike previous approaches, AdaTooler-V adaptively adopts text-based CoT reasoning for problems that do not require tools, while progressively invoking vision tools to refine reasoning for tasks that do.
The core of our approach is a novel reinforcement learning algorithm named Adaptive Tool-use GRPO (AT-GRPO). Specifically, we define a Tool Benefit Score $\Delta S$ for each sample, which quantifies the genuine performance gain provided by tool-use. AT-GRPO adaptively adjusts reward scales based on this score: it rewards tool-use only when it yields tangible improvements and penalizes redundant invocations. 
This mechanism enables the model to autonomously learn a 
favorable and generalizable
reasoning strategy that optimizes both model performance and inference costs.


Besides, to support multimodal joint training, we construct two large-scale datasets: AdaTooler-V-CoT-100k for SFT cold start, and AdaTooler-V-300k for RL training. These datasets cover multiple modalities, including single-image, multi-image, and videos.
They also span diverse visual reasoning tasks such as mathematics, visual counting, logical reasoning, spatial understanding, etc. 
Our two-phase training framework first establishes rich reasoning patterns and behavioral priors during the SFT stage using multi-round tool-interaction trajectories from AdaTooler-V-CoT-100k, and then further optimizes the model’s reasoning strategy in the RL stage using AdaTooler-V-300k combined with the AT-GRPO algorithm. 
This enables AdaTooler-V to perform adaptive tool-use and achieve significant performance improvements over the base model across overall multimodal reasoning benchmarks.
In summary, our contributions are as follows:
\begin{itemize}
    \item We propose \textbf{AdaTooler-V}, an MLLM equipped with adaptive tool-use ability. To support training, we construct two datasets: \textbf{AdaTooler-V-CoT-100k} for SFT and \textbf{AdaTooler-V-300k} for RL training, covering diverse multimodal reasoning tasks and multiple modalities.
    \item We introduce \textbf{AT-GRPO}, a reinforcement learning algorithm that adjusts reward scales using a sample-specific Tool Benefit Score, ensuring tools are invoked only when they provide genuine improvements.
    \item Comprehensive experiments across 12 benchmarks demonstrate the effectiveness of AdaTooler-V. Notably, AdaTooler-V-7B achieves 89.8\% accuracy on V* bench, outperforming the proprietary GPT-4o model.
\end{itemize}

\section{Related Work}
\paragraph{Multimodal Reasoning. }
The field of multimodal large language model reasoning seeks to develop models with human-level capabilities for addressing complex tasks that demand comprehensive understanding and inference across diverse modalities \cite{li2025perceptionreasonthinkplan,wang2025timer1, shen2025vlmr1,zheng2025architecture,liu2025visualrft,wang2025tmcir}.
DeepSeek-R1 \cite{deepseek_r1} has demonstrated that reinforcement learning (RL)-based post-training can significantly enhance the reasoning capabilities of large language models (LLMs). 
Building upon the R1 paradigm, several subsequent works
\cite{ wang2025videorft, yuan2025mme, wang2025knowing,li2025editthinker}
have applied similar post-training paradigms to multimodal large language models (MLLMs) to boost their performance across a variety of tasks. These include:
Mathematical and scientific image-based visual question answering (VQA) \cite{peng2025lmmr1, huang2025visionr1};
Image segmentation and grounding \cite{liu2025seg, bai2025univgr1, shen2025vlmr1, liu2025visualrft, wang2025vgr, wang2025traceable, yang2024lavt}; Video-based VQA \cite{feng2025videor1, wang2025videorft, li2025temporalrlt, cheng2025videoholmes}; Video spatial and temporal grounding \cite{wang2025timer1, li2025videochatr1, park2025deepvideor1, ge2025hunyuanvideo7b}.
Unlike prior approaches that predominantly rely on text-based CoT, we adopt multimodal interleaved CoT, allowing the model to ground intermediate reasoning steps in visual observations and thereby enhance its visual understanding capabilities.

\begin{figure*}[t!]
    \centering
    \includegraphics[width=\linewidth]{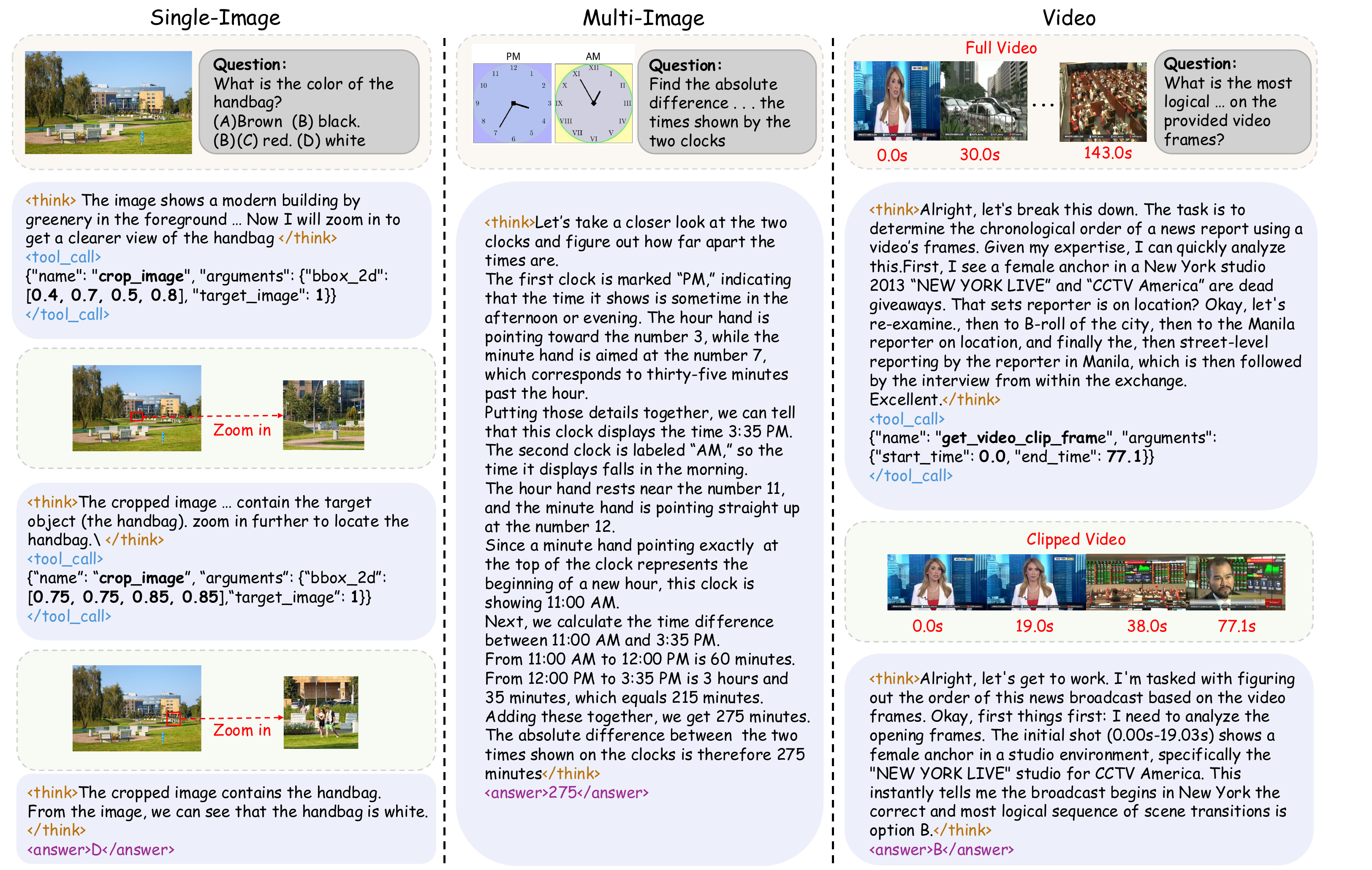}
    \caption{\textbf{Case reasoning trajectory of AdaTooler-V.} For single-image and video questions, the model alternates between internal reasoning, vision tool invocations and final answers, enabling zoom-in on fine-grained regions and inspection of informative clips. In contrast, for the multi-image clock example, AdaTooler-V solves the problem purely via text-based CoT, illustrating its ability to adaptively decide when vision tools are truly necessary.}
    \label{fig:example}
    \vspace{-0.45cm}
\end{figure*}

\paragraph{Thinking with Images. }
The “thinking with images” paradigm has emerged as a promising direction for enhancing multimodal reasoning capabilities. Unlike text-based chain-of-thought (CoT) reasoning, this framework enables models to dynamically invoke operations such as local zooming or video frame extraction, allowing them to progressively explore visual regions, verify hypotheses, and narrow the solution space \cite{pixelreasoner, su2025openthinkimg, ThinkingwithImages, o3, deepeyes, zhang2025thinkingvideosmultimodaltoolaugmented}.
For example, 
OpenThinkIMG \cite{su2025openthinkimglearningthinkimages} introduces an end-to-end visual-tool reinforcement learning framework.
MVoT \cite{li2025imagine} conceptualizes visualization as an intermediate representation within the reasoning process.
PixelReasoner \cite{wang2025pixelreasonerincentivizingpixelspace} leverages curiosity-driven reinforcement learning to incentivize pixel-level reasoning capabilities.
Whereas, VITAL \cite{zhang2025thinkingvideosmultimodaltoolaugmented} explores incorporating multimodal interleaved CoT into video reasoning, thereby enhancing the model’s video comprehension capabilities.
Despite the remarkable progress of these approaches in multimodal reasoning, existing models often exhibit blind tool-use invocation during the reasoning process.


\section{Method}
\subsection{Overview}

\paragraph{Motivation.}
In multimodal reasoning tasks, some problems can be accurately answered through text-based chain-of-thought (CoT) reasoning, while others require tools to perceive visual details. 
However, existing models invoke tools even when they are unnecessary, triggering model “overthinking” and interfering with the correct reasoning path.
Moreover, redundant tool-use incurs additional inference overhead.
Motivated by this observation, we propose a new perspective for multimodal interleaved CoT reasoning: enabling reasoning models to adaptively choose between text-based CoT reasoning and multimodal interleaved CoT reasoning by determining whether a visual problem truly requires tool-use. This approach reduces inference cost while maintaining or even improving overall model performance. To achieve this, we introduce AT-GRPO, a novel reinforcement learning algorithm designed to guide the model in deciding when to invoke tools during the reasoning process.

\begin{figure*}[t!]
    \centering
    \includegraphics[width=0.95\linewidth]{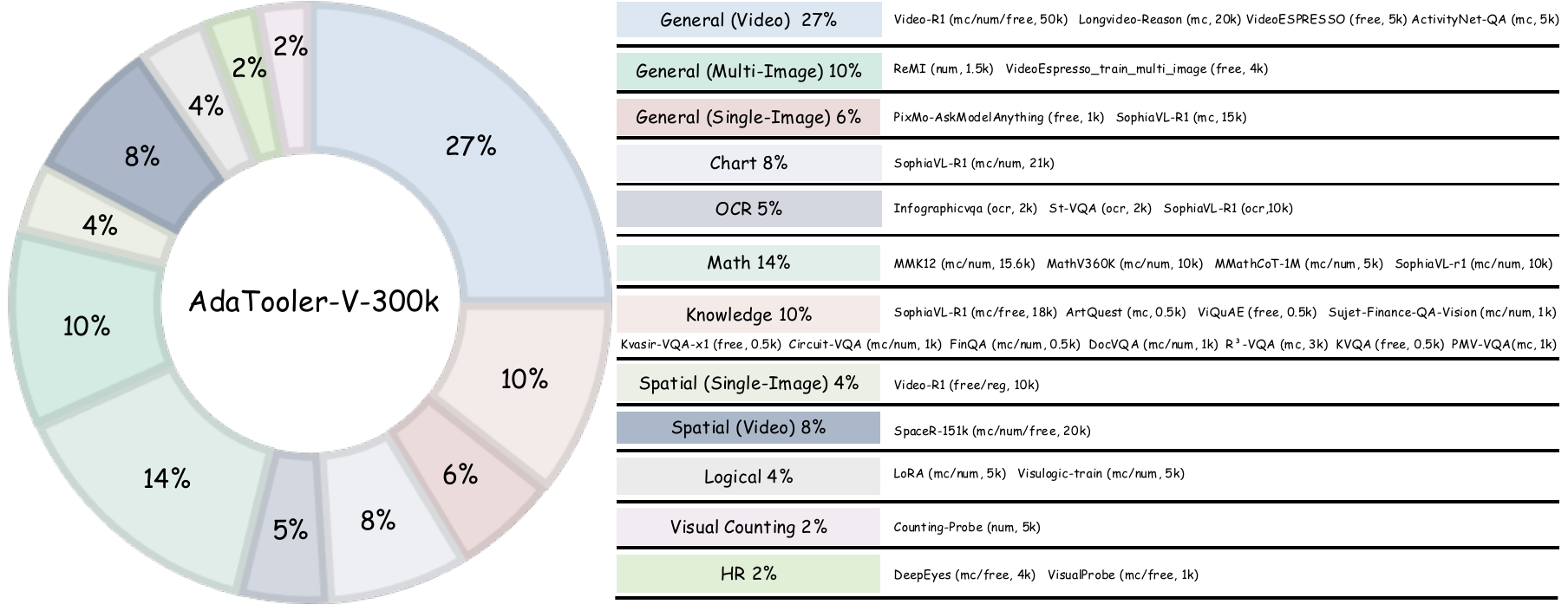}
    \caption{The data distribution of our AdaTooler-V-300k dataset.}
    \label{fig:dataset}
\end{figure*}

\paragraph{Overall Agentic Pipeline.}

Given a user query and an input image/video, the policy model adaptively decides whether to invoke tools.
For problems that don't require tool-use, the model can directly produce a single thought $T$ to derive the final answer.

In contrast, when facing problems that require tool-use, the model enters an iterative reasoning process,
sequentially generating thoughts $T_i$ and actions $C_i$.
Through continuous interaction with the environment, the model progressively refines its understanding until it reaches the final answer.
The action interacts with the environment by invoking image-related tools, yielding a new \emph{observation} $E_i$.
This observation is appended to the interaction history and subsequently fed back into the policy model.
The thought--action--observation loop continues until the model outputs a final answer or reaches a predefined limit on context length or interaction turns.
The core components of the pipeline are detailed below.

\begin{itemize}
\item Thought ($T_i$): Represents the model's internal reasoning process for selecting the next action, conditioned on the accumulated interaction history and the most recent observation.
To enhance exploratory reasoning in complex scenarios, we encourage diverse reasoning trajectories within thoughts to facilitate trial-and-error exploration.

\item Action ($C_i$): The action space includes four primary vision tools: (1) \textit{CropImg: }Zooms in or crops the image based on the specified bounding box. (2) \textit{FrameAt: }Retrieves a single frame from the video at a specific time (in seconds). (3) \textit{VideoClip: }Extracts a video clip between a start and end time; and (4) \textit{PathTracer: }Draws a trajectory or connection between two points on the specified image.
This formulation enables the model to act flexibly on any intermediate observation within the reasoning trajectory.

\item Observation ($E_i$): The visual feedback resulting from executing $C_i$ in the environment. 
Specifically, $E_i$ corresponds to an image patch cropped from either the original input or a historical observation.
\end{itemize}

\paragraph{Two-Phase Training.}
Our training framework consists of two stages:
\begin{itemize}
\item Supervised Fine-Tuning (SFT):
The model is initially fine-tuned on thousands of multi-turn trajectories involving tool interactions, serving as cold-start data.
This stage aims to enable the model to generate coherent trajectories characterized by diverse and robust reasoning patterns.

\item Reinforcement Learning with Verifiable Rewards (RLVR): 
Following SFT, we continue training the model using the proposed AT-GRPO algorithm. This reinforcement learning phase aims to guide the model beyond the rigid pattern-matching behavior established during SFT, encouraging it to autonomously explore more effective reasoning strategies.

\end{itemize}


\subsection{Training Data Collection}

High-quality training data is essential for enhancing visual reasoning capabilities in MLLMs. In this section, we describe the construction of AdaTooler-V-300k for RL training and AdaTooler-V-CoT-100k for SFT cold-start.

\paragraph{Data Collection and Curation.} The dataset aims to cover multiple modalities, including single-image, multi-image, and video. These image-based samples primarily serve to teach the model a broad spectrum of reasoning skills across various domains and difficulty levels, including mathematics, spatial logic, and expert-level knowledge. Such data enable the model to acquire generalized reasoning capabilities in static contexts. In contrast, video-based data are utilized to enhance the model’s temporal reasoning ability, allowing it to understand event progression, capture frame-to-frame dependencies, and infer outcomes based on motion and causal dynamics over time.

Our dataset is constructed from multiple public sources, with careful sampling and balancing across different subsets. The final composition of AdaTooler-V-300k is summarized in Fig. \ref{fig:dataset} and additional details are provided in Appendix \ref{data_dis}.
\begin{figure*}[t!]
    \centering
    \includegraphics[width=\linewidth]{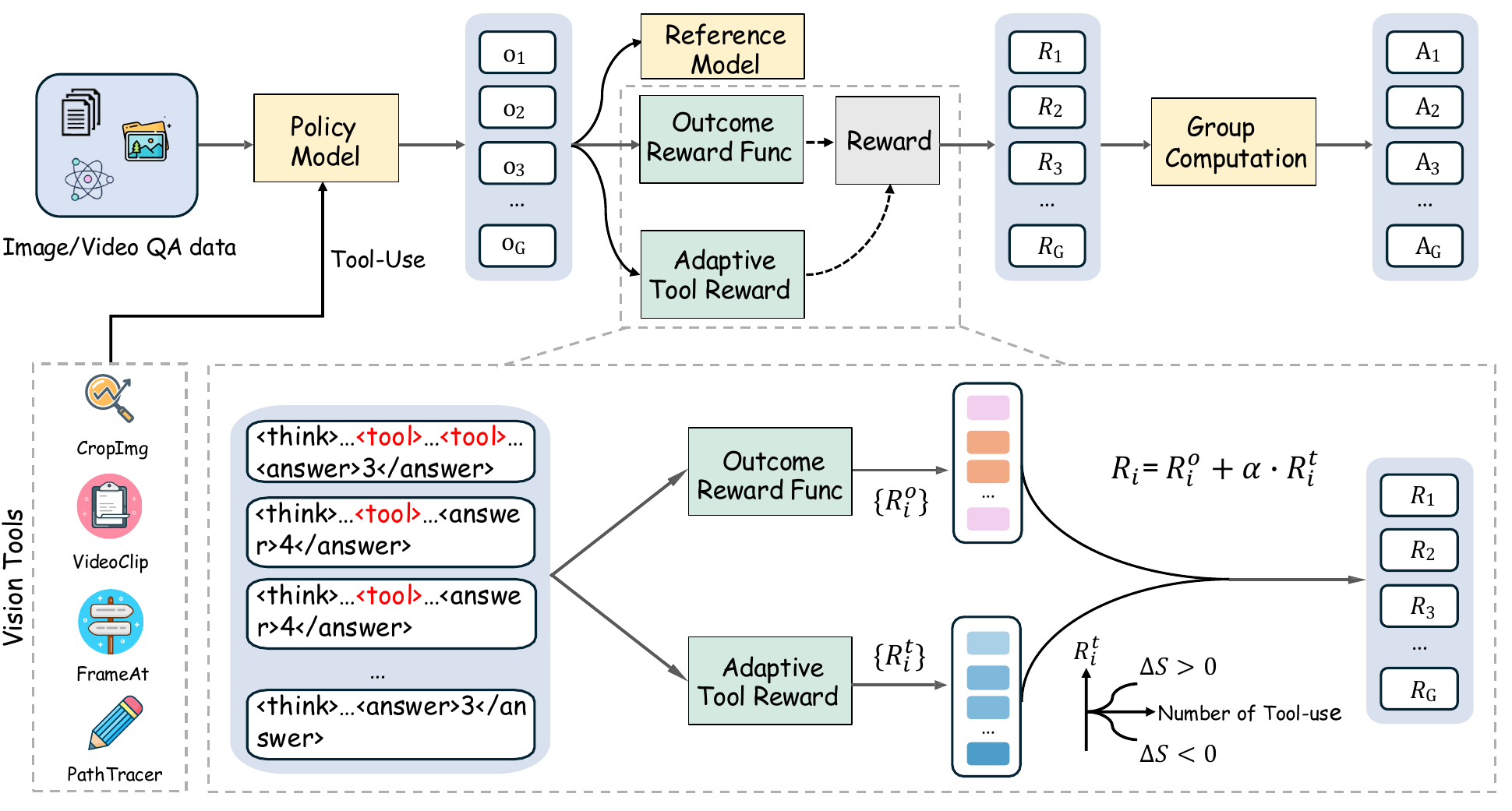}
\caption{An illustration of our proposed AT-GRPO.}
    \label{fig:atgrpo}
    \vspace{-4mm}
\end{figure*}


  \vspace{-0.1in}

\paragraph{CoT Annotation.} 


To facilitate effective initialization during the SFT stage, we leverage Qwen2.5-VL-72B-Instruct \cite{bai2025qwen2} to automatically produce Chain-of-Thought (CoT) rationales for all samples in AdaTooler-V-300k. The complete prompt specification employed for CoT generation is included in Appendix \ref{prompt_template}. Following generation, we apply a sequence of rule-based filtering procedures to eliminate low-quality or semantically inconsistent outputs. This process yields a high-fidelity corpus, AdaTooler-V-CoT-100k, which forms the foundation for the cold-start stage of SFT.

\paragraph{Data Type and Rule-based Reward Design.}

Our RL framework follows the rule-based reward paradigm of DeepSeek-R1 \cite{deepseek_r1}, necessitating reward signals that are both reliable and precise. To this end, the majority of training samples are designed around tasks with easily verifiable outputs, such as multiple-choice or numerical answer formats, enabling accurate reward computation via simple rules and ensuring stable RL training. 

To improve the model’s flexibility and generalization across diverse tasks and formats, we additionally incorporate a smaller portion of more complex data types, including free-form generation, OCR tasks, and regression problems, which are critical for real-world applications. 

The data types and their corresponding reward functions are summarized as follows:

\begin{itemize}
\item \textbf{Multiple Choice:} Rewards are assigned based on an exact match between the model prediction and the ground-truth option.
\item \textbf{Numerical QA:} Rewards are given according to whether the predicted numerical value precisely matches the reference.
\item \textbf{OCR:} Rewards are computed using the Word Error Rate (WER), which measures the edit distance between the predicted text and the ground-truth transcription.
\item \textbf{Free-form QA:} Rewards are determined by the average of ROUGE-1, ROUGE-2, and ROUGE-L scores, assessing the similarity between the generated output and the reference answer.
\end{itemize}

\begin{table*}[]
\centering
\small
\vspace{12pt}
\caption{Comparison of models on single-image and multi-image benchmarks. The first six evaluation benchmarks belong to single-image comprehension tasks, and the last two evaluation benchmarks belong to multi-image understanding tasks. }
\label{tab:general_results}
 \renewcommand\arraystretch{1.1}
\setlength{\tabcolsep}{4pt}
\begin{tabular}{@{}lccccccc@{}}
\toprule
\textbf{Model} & \textbf{V*} & \textbf{MME} & \textbf{InfoVQA} & \textbf{MMBench} & \textbf{MathVista} & \textbf{MMSI-Bench} & \textbf{SPAR-Bench} \\
\midrule
\rowcolor[HTML]{F2F2F2} \multicolumn{8}{l}{\textit{Proprietary Models}} \\
\midrule
GPT-4o~\cite{openai_gpt4o_2024} & 65.2 & 2328 & 80.7 & 82.1 & 63.8 & 30.3 & 33.6 \\
Gemini 1.5 Pro \cite{gemini15pro_2024} & 71.7 & -- & 81.0 & -- & 63.9 & 36.9 & -- \\
\midrule
\rowcolor[HTML]{F2F2F2} \multicolumn{8}{l}{\textit{Open-Source Models}} \\
\midrule
InternVL3-8B~\cite{zhu2025internvl3exploringadvancedtraining} & -- & 2415.4 & 76.8 & 83.4 & 71.6 & 25.7 & -- \\
LLaVA-1.5-7B~\cite{liu2024improved} & -- & 1510.7 & -- & 64.3 & -- & -- & 23.65 \\
LLaVA-OneVision-7B~\cite{li2024llava} & -- & 1580.0 & 68.8 & 80.8 & 63.2 & -- & -- \\
SophiaVL-R1-7B \cite{fan2025sophiavl} & -- & 2403.8 & -- & 85.4 & 71.3 & -- & -- \\
Qwen2.5-VL-7B-Instruct~\cite{bai2025qwen25vltechnicalreport} & 78.5 & 2347.0 & 82.6 & 83.4 & 68.2 & 25.9 & 33.07 \\  \midrule
\rowcolor[HTML]{F2F2F2} \multicolumn{8}{l}{\textit{Open-Source o3-like Image Models}} \\
\midrule
Pixel Reasoner~\cite{wang2025pixelreasonerincentivizingpixelspace} & 84.3 & -- & 84.0 & -- & -- & -- & -- \\
DeepEyes~\cite{zheng2025deepeyes} & 85.6 & -- & -- & -- & 70.1 & -- & -- \\
Mini-o3~\cite{lai2025mini} & 88.2 & -- & -- & -- & -- & -- & -- \\
Thymes \cite{zhang2025thyme} & 82.2 & -- & -- & -- & 70.0 & -- & --  \\
VILASR \cite{wu2025reinforcing} & -- & -- & -- & -- & -- & 30.2 & 37.6  \\
\midrule
\rowcolor[HTML]{DAEFF9}
AdaTooler-V-7B & \textbf{89.8} & \textbf{2460.8} & \textbf{86.0} & \textbf{87.8} & \textbf{74.5} & \textbf{36.8} & \textbf{40.3} \\
\bottomrule%
\end{tabular}
\label{tab:main_results_image}
\end{table*}

\subsection{Adaptive Tool-use GRPO Training}

To enable adaptive tool-use during the reasoning process, we propose Adaptive Tool-use GRPO (AT-GRPO), which guides the model to invoke tools only when they offer a  genuine performance gain, thereby improving model performance and reducing inference overhead, as illustrated in Fig. \ref{fig:atgrpo}.


For each input query $q_i$, we define a Tool Benefit Score $\Delta S_i$ during the data annotation stage to quantify the performance improvement brought by tool-use:
\begin{equation}
\Delta S_i = S^+(q_i)-S^-(q_i)
\end{equation}
Here, $S^+$ and $S^-$ denote the average accuracy of the model when reasoning with and without tool-use, respectively. 
Each query is evaluated eight times using Qwen2.5-VL-72B-Instruct \cite{bai2025qwen2} (i.e., eight runs with tool-use and eight runs without tool-use)
and the averaged accuracy gap is used as $\Delta S_i$.

The adaptive-tool reward $R_i^t$ is then formulated as:

\begin{equation}
R^t_i = \Delta S_i \cdot \exp\left(-\gamma\left(\frac{n_{\text{tool}}-n_{\text{max}}}{n_{\text{max}}}\right)^2\right)
\end{equation}
where $\Delta S_i$ represents the improvement in accuracy attributed to tool-use, $n_{\text{tool}}$ is the number of tool-use during the reasoning trajectory,  $n_{\text{max}}$ denotes the maximum allowable number of tool-use, and $\gamma$ controls the sensitivity of the Gaussian decay to tool frequency, making the reward variation smoother. Here we set $\gamma$ = 2.

This design encourages the model to adaptively invoke tools by determining whether a visual problem truly requires tool-use.
Specifically, when a problem does not warrant the use of vision tools ($\Delta S_i < 0$), the model is penalized if it still invokes such tools during the reasoning process, and the penalty increases progressively with the number of tool-use. In contrast, when a problem benefits from tool-use ($\Delta S_i > 0$), the model receives a positive reward and the reward similarly grows with the number of tool-use.
We adopt this exponential form for its simplicity, differentiability, and stable gradient behavior during training.

The total reward for response $i$ is defined as:
\begin{equation}
\label{final_reward}
R_i = R_i^o + \alpha  \cdot R_i^t
\end{equation}
where $\alpha$ balances the relative weight of tool-use reward in the total objective and
$R_i^o$ denotes the base reward of response $i$, including correctness and formatting components, following the formulation in~\cite{guo2025deepseek}. The combined reward $R_i$ is then used to compute the advantage for policy optimization during GRPO training. 

\begin{table*}[]
\centering
\small
\vspace{12pt}
\caption{Comparison of models on video benchmarks.}
 \renewcommand\arraystretch{1.1}
\label{tab:general_results_video}
\setlength{\tabcolsep}{4pt}
\begin{tabular}{@{}lcccccc@{}}
\toprule
\textbf{Model} & \textbf{Frames} & \textbf{VSI-Bench} & \textbf{VideoMMMU} & \textbf{MVBench} & \textbf{Video-MME(w/o sub)} & \textbf{Video-Holmes} \\
\midrule
\rowcolor[HTML]{F2F2F2} \multicolumn{7}{l}{\textit{Proprietary Models}} \\
\midrule
GPT-4o~\cite{openai_gpt4o_2024} & -- & 34.0&  61.2& 64.6& 71.9& 42.0\\
Gemini 1.5 Pro~\cite{gemini15pro_2024} & -- & 45.4& 53.9& 60.5& 75.0& 41.2\\
\midrule
\rowcolor[HTML]{F2F2F2} \multicolumn{7}{l}{\textit{Open-Source Models}} \\
\midrule
InternVL3-8B~\cite{zhu2025internvl3exploringadvancedtraining} & -- & 42.1 & -- & 75.4 & 66.3& --\\
VideoChat-R1 \citep{li2025videochat} & -- & --& --& 67.9 & 72.2& 33.0\\
Video-CCAM \citep{fei2024videoccame}  & -- & --& --& 62.8& 50.1& --\\
Video-XL \citep{video_xl}  & -- & --& 52.3& 55.3& 55.5& --\\
Qwen2.5-VL-7B-Instruct \citep{bai2025qwen25vltechnicalreport} & 32 & 29.8& 47.4& 58.2& 56.1& 27.8\\
Qwen2.5-VL-7B-Instruct \citep{bai2025qwen25vltechnicalreport} & 64 & 30.9& 49.1& 59.8& 58.6& 29.9\\
Qwen2.5-VL-7B-Instruct \citep{bai2025qwen25vltechnicalreport} & 128 & 34.8& 51.3& 62.3& 60.4& 33.5\\
Video-R1 \citep{feng2025videor1} & -- & 37.1& 52.4& 64.8& 61.4& 36.5\\
\midrule
\rowcolor[HTML]{F2F2F2} \multicolumn{7}{l}{\textit{Open-Source o3-like Video Models}} \\
\midrule
FrameMind \cite{ge2025framemind}  & -- & --& --& 64.2 & 60.9 & --\\
Open-o3 Video \cite{meng2025open}  & -- & --& 52.3 & -- & 63.6 & --\\
Video-Thinker \cite{wang2025video} & -- & --& --& -- & -- & 43.2\\
VILASR \cite{wu2025reinforcing} & -- & 45.4& --& -- & -- & --\\
\midrule
\rowcolor[HTML]{DAEFF9}
AdaTooler-V-7B & 32 & 46.7 & 54.6 & 68.4 & 62.5 & 55.6 \\
\rowcolor[HTML]{DAEFF9}
AdaTooler-V-7B & 64 & 47.9 & 55.1 & 70.2 & 63.4 & 56.4 \\
\rowcolor[HTML]{DAEFF9}
AdaTooler-V-7B & 128 & \textbf{49.5} & \textbf{56.8} & \textbf{71.5} & \textbf{66.7} & \textbf{58.3} \\
\bottomrule
\end{tabular}
\label{tab : main_results_video}
\end{table*}

The advantage $A_i$ is calculated within each group as:
\begin{equation}
A_i = \frac{R_i - \text{mean}(\{R_1, R_2, \dots, R_G\})}{\text{std}(\{R_1, R_2, \dots, R_G\})}
\end{equation}
Following DeepSeek-R1~\cite{deepseek_r1}, the final policy objective for AT-GRPO is given by:
\begin{align}
\mathcal{J}_{\mathrm{AT-GRPO}}(\theta) &=
\mathbb{E}_{q \sim P(Q),\ \{o_i\}_{i=1}^G \sim \pi_{\theta_{\text{old}}}(o|q)} \notag \\
\bigg[
\frac{1}{G} \sum_{i=1}^G &
\frac{\pi_{\theta}(o_i|q)}{\pi_{\theta_{\text{old}}}(o_i|q)} A_i
- \beta\, \mathbb{D}_{\mathrm{KL}}(\pi_{\theta} \| \pi_{\mathrm{ref}})
\bigg]
\end{align}
Through this formulation, AT-GRPO enables the model to autonomously learn to adaptively invoke vision tools, thereby enhancing model performance and reducing inference cost.

\section{Experiments}
\subsection{Setup}
\label{experimental_setup}
\paragraph{Benchmarks.}
Following prior works \cite{feng2025videor1, xiao2025proxythinkertesttimeguidancesmall,wang2025vlrethinkerincentivizingselfreflectionvisionlanguage},
we employ greedy decoding to systematically evaluate our proposed model and other baselines across a diverse suite of multimodal benchmarks, encompassing critical capabilities such as general knowledge comprehension, high-resolution image detail perception, logical reasoning, spatial reasoning, and chart interpretation. Specifically, for the image modality, we select seven representative benchmarks: V* \cite{wang2023vstarvideogroundeddialoguedataset}, MME \cite{fu2025mmecomprehensiveevaluationbenchmark}, InfoVQA \cite{mathew2021infographicvqa}, MMBench \cite{liu2024mmbenchmultimodalmodelallaround}, MathVista \cite{lu2024mathvistaevaluatingmathematicalreasoning}, MMSI-Bench \cite{yang2025mmsibenchbenchmarkmultiimagespatial}, and SPAR-Bench \cite{zhang2025flatlandspaceteachingvisionlanguage}. For the video modality, we adopt five representative benchmarks: VSI-Bench \cite{yang2025thinkingspacemultimodallarge}, VideoMMMU \cite{hu2025videommmuevaluatingknowledgeacquisition}, MVBench \cite{li2024mvbenchcomprehensivemultimodalvideo}, Video-MME \cite{fu2025videommefirstevercomprehensiveevaluation}, and Video-Holmes \cite{cheng2025videoholmesmllmthinklike}.

\paragraph{Implementation Details.}
We use 8 NVIDIA H100 (80GB) GPUs to train our model. The training framework is based on verl-tool \cite{jiang2025verltool}, which extends the functionalities of verl \cite{sheng2024hybridflow} and vLLM \cite{kwon2023efficient}, providing additional support for multimodal tool-augmented multi-turn training and evaluation.
Our model is initialized based on Qwen2.5-VL-7B-Instruct \cite{qwen2.5-VL}. 
First, we perform supervised fine-tuning (SFT) on the AdaTooler-V-CoT-100k dataset to obtain the Qwen2.5-VL-7B-SFT model, where the number of epochs is set to 1, the batch size is set to 16, and the learning rate is set to 5.
Subsequently, we conduct reinforcement learning (RL) training on the AdaTooler-V-300k dataset to generate the final AdaTooler-V model, where the batch size is set to 32, the KL divergence coefficient to 0.04, and 
the learning rate to $5 \times 10^{-7}$. 
The maximum response length is limited to 4096 tokens.
The model is optimized with AdamW \cite{loshchilov2019decoupledweightdecayregularization} throughout the training process.
The hyperparameter $\alpha$ in Eqn. \ref{final_reward} is set to 0.6.

\subsection{Main Results}

\paragraph{Image Benchmarks.}
As summarized in Tab.\ref{tab:main_results_image}, AdaTooler-V-7B achieves state-of-the-art performance across multiple single-image comprehension datasets. On the challenging high-resolution V* benchmark, our method reaches 89.8\% accuracy, outperforming recent tool-based models including Pixel Reasoner (84.3\%), DeepEyes (85.6\%), and Mini-o3 (88.2\%). Moreover, compared to the base model Qwen2.5-VL-7B-Instruct, AdaTooler-V provides a substantial +11.3\% absolute improvement, highlighting the effectiveness of adaptive tool-use in high-resolution visual reasoning.
Beyond high-resolution tasks, AdaTooler-V also shows consistent gains on general reasoning benchmarks such as MME, MathVista, InfoVQA, and MMBench, indicating strong cross-domain generalization. Notably, the model achieves 74.5\% on MathVista, surpassing Qwen2.5-VL-7B-Instruct by over 6 percentage points. These improvements suggest that controlling tool-invocation frequency allows the model to focus on essential visual cues rather than over-processing evidence.
On multi-image reasoning tasks, AdaTooler-V-7B yields clear advantages on MMSI-Bench (36.8) and SPAR-Bench (40.3), outperforming all tested baselines. Since these benchmarks require spatial correspondence and relational reasoning across multiple images, results reflect AdaTooler-V’s ability to selectively invoke tools when information extraction becomes visually non-trivial.

\paragraph{Video Benchmarks.}
As is illustrated in Tab. \ref{tab : main_results_video}, AdaTooler-V displays substantial performance gains over strong video-reasoning baselines. For example, our model achieves 46.7\% on VSI-Bench, 54.6\% on VideoMMMU, and 68.4\% on MVBench using only 32 frames, surpassing both Qwen2.5-VL-7B-Instruct and Video-R1 based models.
The Video-Holmes benchmark further highlights AdaTooler-V’s strengths in complex, long-range video reasoning. Our method obtains 55.6\%, compared to 27.8\% for Qwen2.5-VL-7B-Instruct and 36.5\% for Video-R1, showing more than a 2× improvement over the base model in causal, sequential inference settings.
Moreover, we observe consistent performance gains across nearly all benchmarks as the number of input frames increases. This suggests that richer contextual cues and temporal information can further enhance the model’s reasoning capability.

\begin{table}[t]
    \caption{Ablation study on training stages.}
    \small
    \resizebox{\linewidth}{!}{%
    \setlength{\tabcolsep}{0.8mm}
    \renewcommand\arraystretch{1.3}

    \begin{tabular}{l l cccc c}
        \Xhline{0.8pt}
        & \textbf{Train Stage} & \textbf{V*} & \textbf{MathVista} & \textbf{VSI-Bench} & \textbf{MVBench} & \textbf{Avg.} \\
        \hline
        & GRPO               & 85.1 & 71.8 & 40.7 & 65.9 & 65.9 \\
        & SFT+GRPO           & 87.0 & 73.2 & 42.3 & 67.7 & 67.6 \\
        & SFT+AT-GRPO        & 89.8 & 74.5 & 46.7 & 68.4 & 69.9 \\
        \Xhline{0.8pt}
    \end{tabular}
    \label{tab:abl_train}
    }
\end{table}

\begin{table}[t]
\centering

\caption{Ablation study on the $\alpha$ in Eqn. \ref{final_reward}.}
\small
    \resizebox{\linewidth}{!}{%
    \setlength{\tabcolsep}{3mm}
    \renewcommand\arraystretch{1.1}
\begin{tabular}{ccccc c}
\toprule
\textbf{$\alpha$} & \textbf{V*} & \textbf{MathVista} & \textbf{VSI-Bench} & \textbf{MVBench} & \textbf{Avg.} \\
\midrule
0.2 & 88.1 & 73.6 & 44.2 & 67.9 & 68.5 \\
0.4 & 88.9 & 74.1 & 43.9 & 68.2 & 68.7 \\
0.6 & 89.8 & 74.5 & 46.7 & 68.4 & 69.9 \\
0.8 & 89.2 & 73.9 & 45.1 & 68.1 & 69.1 \\
\bottomrule
\end{tabular}
}
\label{tab:alpha_ablation}
\end{table}

\begin{table}[t]
    \caption{Ablation study on tool-use.}
    \resizebox{\linewidth}{!}{%
    \setlength{\tabcolsep}{0.65mm}
    \renewcommand\arraystretch{1.3}

    \begin{tabular}{l l cccc c}
        \Xhline{0.8pt}
        & \textbf{Model} & \textbf{V*} & \textbf{MathVista} & \textbf{VSI-Bench} & \textbf{MVBench} & \textbf{Avg.} \\
        \hline
        & Qwen2.5-VL-7B            & 78.5 & 68.2 & 31.8 & 63.8 & 60.6 \\
        & RL-wo-tool     & 84.4 & 72.6 & 39.9 & 65.0 & 65.5 \\
        & AdaTooler-V-7B     & 89.8 & 74.5 & 46.7 & 68.4 & 69.9 \\
        \Xhline{0.8pt}
    \end{tabular}
    \label{tab:abl_tool}
    }
\end{table}


\subsection{Ablation Study}

\subsubsection{Effectiveness of AT-GRPO}
To validate the effectiveness of the proposed AT-GRPO algorithm, we compare multiple training stages: vanilla GRPO, which skips the SFT cold-start phase and directly applies reinforcement learning for training; SFT+GRPO, which replaces our proposed AT-GRPO algorithm with the standard GRPO method.
As shown in the last two rows of Tab. \ref{tab:abl_train}, incorporating the proposed AT-GRPO training strategy leads to a substantial performance improvement. These results confirm that dynamically adjusting tool-use rewards based on the Tool Benefit Score enables the model to invoke tools only when necessary, effectively avoiding redundant visual interactions and leading to more accurate reasoning.

\subsubsection{Necessity of the SFT}
We further investigate the necessity of the supervised fine-tuning (SFT) stage prior to reinforcement learning. As shown in the first row of Tab. \ref{tab:abl_train}, skipping the SFT cold start also leads to degraded performance. This is primarily because the model lacks structured priors for tool-interaction reasoning, making it difficult to produce coherent and well-formed reasoning trajectories during the early stages of RL training.
In contrast, introducing SFT as a cold start equips the model with essential tool-use patterns and multimodal reasoning priors, which serve as a strong foundation for subsequent RL optimization. This initialization leads to a more stable and progressive training process, enabling the model to better exploit the benefits of reinforcement learning.
Overall, SFT plays a critical role in bootstrapping structured tool-use behaviors, and is indispensable for enabling RL to further refine adaptive reasoning strategies.

\subsubsection{Analysis of $\alpha$}
We further perform a analysis on the magnitude of the adaptive-tool reward, governed by the hyperparameter $\alpha$, as illustrated in Tab. \ref{tab:alpha_ablation}. We observe a moderate decrease in performance when $\alpha$ is set to 0.2 or 0.4, whereas $\alpha$ values of 0.6 and 0.8 yield comparable and consistently strong results. These findings suggest that the model exhibits a low sensitivity to the selection of $\alpha$ within a reasonable range.

\begin{figure}[]
    \centering
    \begin{subfigure}{0.48\linewidth}
        \centering
        \includegraphics[width=\linewidth]{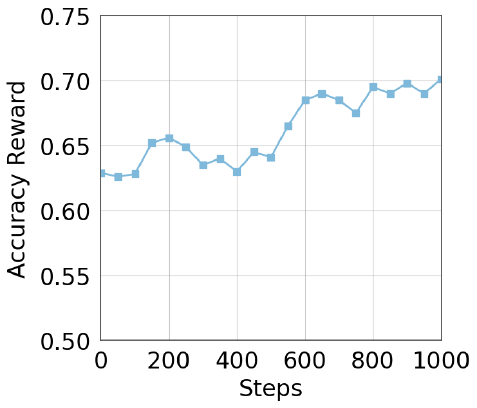}
        \caption{Accuracy Reward}
        \label{fig:alpha_a}
    \end{subfigure}
    \hfill
    \begin{subfigure}{0.48\linewidth}
        \centering
        \includegraphics[width=\linewidth]{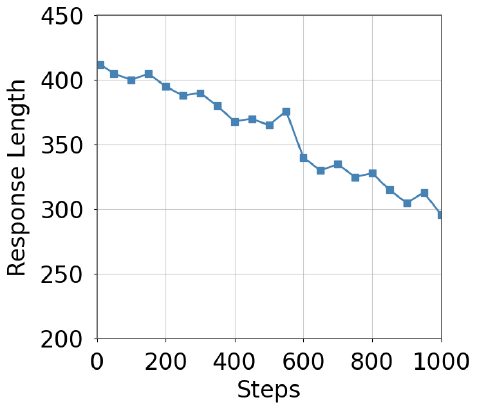}
        \caption{Response Length}
        \label{fig:alpha_b}
    \end{subfigure}
    \caption{
    RL training curves.
    }
    \label{fig:tra_curves}
\end{figure}

     


\subsubsection{Effectiveness of Tool-use}
To assess the effectiveness of tool-use, we train a variant of Qwen2.5-VL-7B-Instruct using end-to-end RL with text-based CoT reasoning on the same training dataset. 
As shown in Tab. \ref{tab:abl_tool}, disabling tool interactions leads to consistent drops across four benchmarks. For example, from 89.8\% to 84.4\% on V* and from 46.7\% to 39.9\% on VSI-Bench. These results verify that vision tool-use provides complementary evidence beyond text-based reasoning and is essential for accurate multimodal understanding.

\subsection{Training Curves}
Fig. \ref{fig:tra_curves} illustrates the dynamics of several critical metrics during the RL process, from which we derive the following observations and findings.
As shown in Fig. \ref{fig:tra_curves} (a), the model's accuracy improves significantly throughout the training process, increasing from approximately 0.60 to around 0.70. This indicates that as the model learns through reinforcement learning and tool interactions, its ability to generate accurate answers continuously improves.
In Fig. \ref{fig:tra_curves} (b), the average response length rapidly decreases in the initial stages before stabilizing. This phenomenon originates from the SFT phase, where the model learns to invoke tools based on instructions to solve problems. However, many of the images in the dataset are relatively simple, making many of the tool-use unnecessary. During the reinforcement learning phase, the model quickly realizes that, for most tasks, generating a direct text response is more efficient than performing a tool-based analysis. As a result, unnecessary tool-use decrease rapidly, leading to a reduction in response length.


\section{Conclusion}

We introduced AdaTooler-V, a multimodal large language model equipped with adaptive tool-use capability.
To achieve this, we introduced AT-GRPO, a reinforcement learning algorithm that leverages a sample-specific Tool Benefit Score to dynamically modulate rewards, encouraging tools to be used only when they provide genuine performance gains.
To support training, we curate two datasets,
AdaTooler-V-CoT-100k for SFT cold start and AdaTooler-V-300k for RL.
Experiments across twelve benchmarks validate the effectiveness of our approach.
We believe it provides a promising foundation for future research on tool-augmented MLLMs.

{
    \small
    \bibliographystyle{ieeenat_fullname}
    \bibliography{main}
}

\newpage
\appendix
\clearpage
\setcounter{page}{1}


\section{Dataset Distribution Details}
\label{data_dis}
The distribution of AdaTooler-V-300k dataset can be roughly categorized as follows:

\begin{itemize}
    \item \textbf{General (Video, 81k):} A diverse collection of open-domain videos depicting everyday scenarios, designed to cultivate temporal comprehension and reasoning.

    \item \textbf{General (Multi-Image, 33k):} Multi-image reasoning tasks that test cross-view comparison and contextual integration.
    
    \item \textbf{General (Image, 18k):} General-purpose image question-answering data for foundational visual understanding.
    
    \item \textbf{Chart (Image, 24k):} Visual reasoning over charts, line graphs, and scientific figures, emphasizing data interpretation and quantitative logic.
    
    \item \textbf{OCR (Image, 15k):} Tasks requiring recognition and interpretation of embedded textual content, such as signs, forms, or documents.
    
    \item \textbf{Math (Image, 42k):} Image-based math reasoning problems, including formulas, geometric diagrams, and multi-step symbolic reasoning.
    
    \item \textbf{Knowledge (Image, 30k):} Visual commonsense and cross-disciplinary reasoning tasks to evaluate the integration of world knowledge with visual cues.
    
    \item \textbf{Spatial (Image, 12k):} Static spatial reasoning such as occlusion and positional inference.

    \item \textbf{Spatial (Video, 24k):} Focused on spatial reasoning in motion, including navigation, object tracking, and path planning, enhancing spatiotemporal understanding.

    \item \textbf{Logical (Image, 12k):} Visual logic tasks involving pattern recognition and rule-based reasoning.

    \item \textbf{Visual Counting (Image, 6k):} Object counting and density estimation for quantitative perception.

    \item \textbf{High-Resolution (Image, 6k):} Fine-grained visual understanding with small-object and texture recognition.
\end{itemize}

\begin{figure*}[h]
    \centering
    \includegraphics[width=1\linewidth]{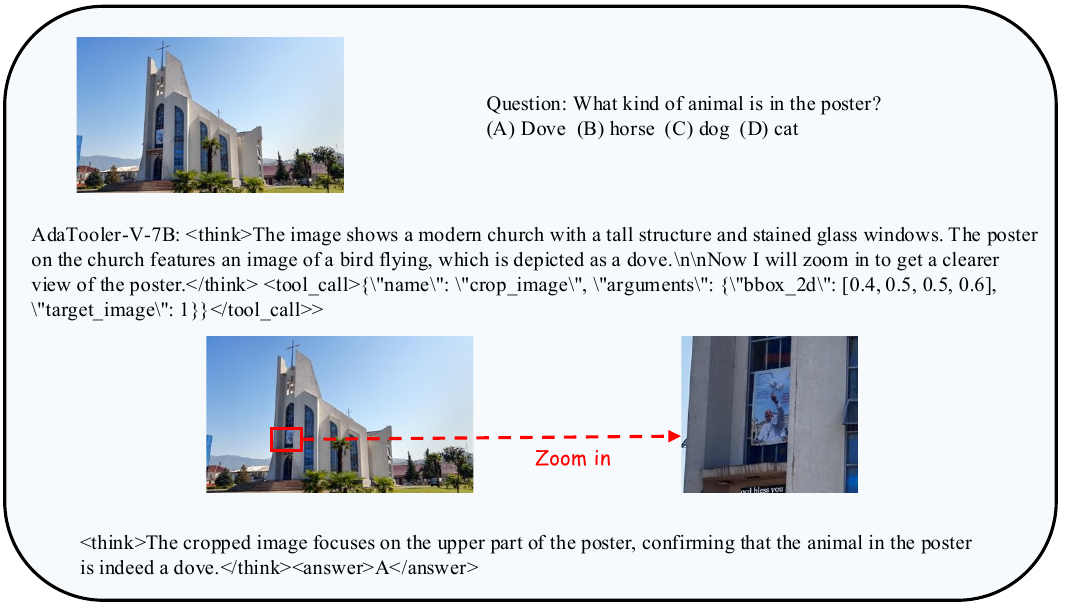}
    \caption{An example of AdaTooler-V-7B’s reasoning output on V* Benchmark.}
    \label{fig:example_1}
\end{figure*}
\begin{figure*}[h]
    \centering
    \includegraphics[width=1\linewidth]{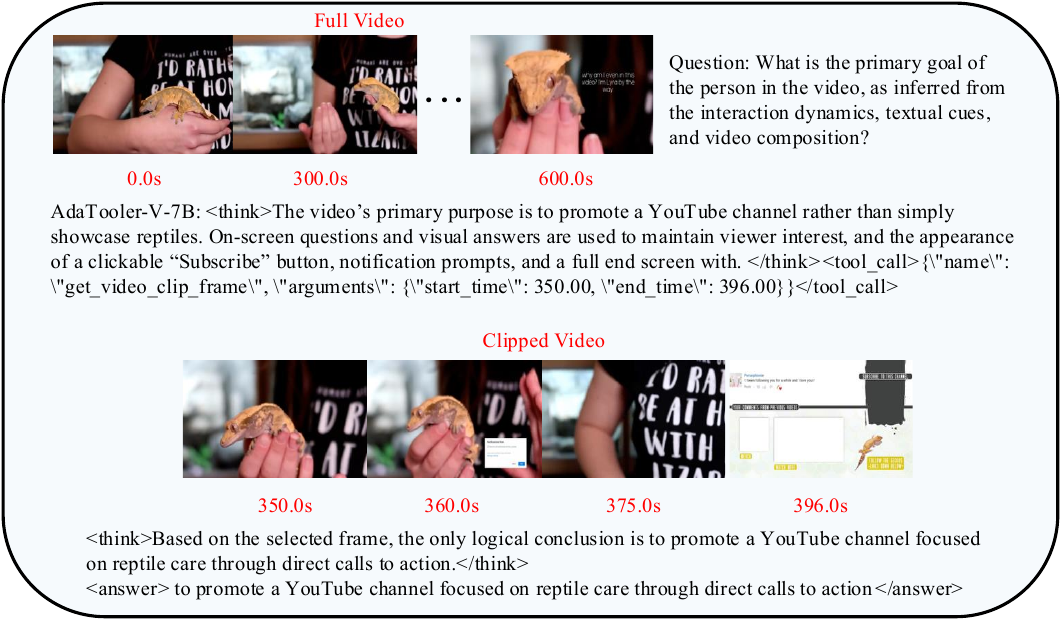}
    \caption{An example of AdaTooler-V-7B’s reasoning output on MVBench.}
    \label{fig:example_2}
\end{figure*}

\begin{figure*}[h]
    \centering
    \includegraphics[width=1\linewidth]{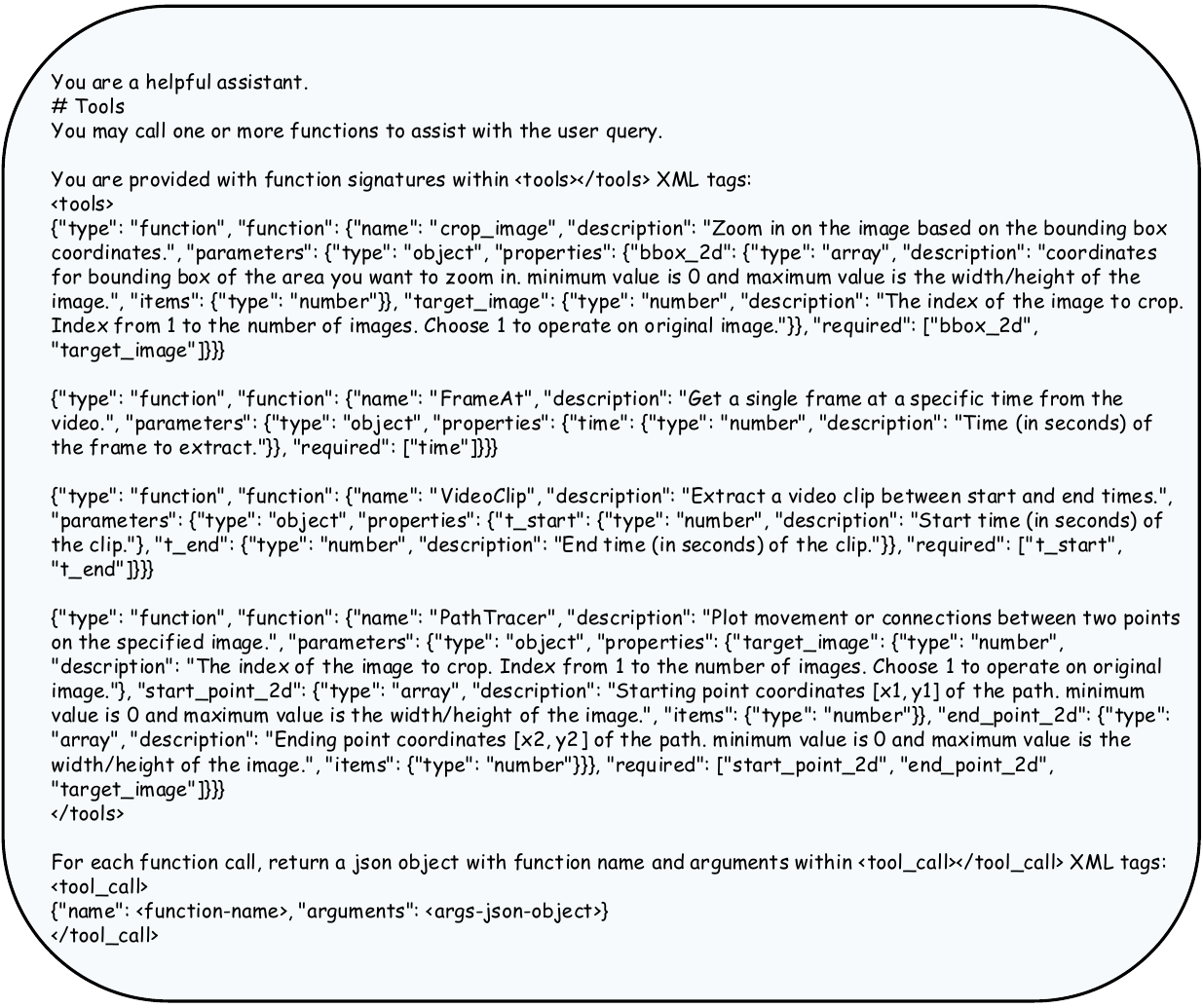}
    \caption{Prompt template for training and inference.}
    \label{fig:prompt_1}
\end{figure*}

\section{Reasoning Examples}
This section presents representative reasoning examples generated by AdaTooler-V-7B, as shown in Fig. \ref{fig:example_1} and Fig. \ref{fig:example_2}.

\section{Prompt Template for Training and Inference}
\label{prompt_template}
Fig. \ref{fig:prompt_1} illustrates the prompt template for training and inference of all models. We also use this prompt for the COT annotation.

\section{Limitations and Future Works}
We outline the limitations of our work and potential avenues for future research as follows:

First, the estimation of tool benefit ($\Delta S$) relies on a single reference model, which may introduce biased assessments of when tool use is genuinely helpful; future work may develop a learned benefit estimator or leverage model ensembles to obtain more robust $\Delta S$ predictions. Second, our reward design primarily targets verifiable tasks such as multiple-choice and numerical reasoning, making it less suitable for open-ended generation; future research could incorporate learned reward models, multimodal discriminators, or contrastive signals to better support free-form reasoning. Third, the AdaTooler-V-300k dataset is mainly constructed from public benchmarks, resulting in limited coverage of real-world long-tail cases, noisy conditions, and cross-domain scenarios; expanding the dataset with in-the-wild samples, hard-case synthesis, or domain adaptation techniques may enhance generalization.

\end{document}